\documentclass[11pt,a4paper]{article}
\usepackage[hyperref]{acl2020}
\usepackage{times}
\usepackage{latexsym}
\usepackage{url}
\usepackage{examples}
\usepackage{multirow}
\usepackage{graphicx}
\usepackage{amsmath}
\usepackage[para,online,flushleft]{threeparttable}
\usepackage{tablefootnote}

\usepackage{microtype}

\aclfinalcopy 
\def\aclpaperid{204} 

\title{
Bridging Anaphora Resolution as Question Answering
}

\author{
Yufang Hou \\
  IBM Research Europe\\ {\tt yhou@ie.ibm.com}\\
   }

\date{}
\begin{document}
\maketitle

\begin{abstract}
Most previous studies on bridging anaphora resolution \cite{poesio04d, houyufang13a, houyufang18b} use the pairwise model to tackle the problem and assume that the gold mention information is given. In this paper, we cast bridging anaphora resolution
as question answering based on context. This allows us to find the antecedent for a given anaphor without knowing any gold mention information (except the anaphor itself). We present a question answering framework (\emph{BARQA})  for this task, which leverages the power of transfer learning. Furthermore, we 
propose a novel method to generate a large amount of ``quasi-bridging'' training data. We show that our model pre-trained on this dataset  and fine-tuned on a small amount of in-domain dataset achieves new state-of-the-art results for bridging anaphora resolution on two bridging corpora (ISNotes \cite{markert12} and BASHI \cite{roesiger18b}).

\end{abstract}

\section{Introduction}
\label{sec:intro}


Anaphora accounts for text cohesion and is crucial for text understanding. 
 An anaphor is a noun phrase (NP) that usually refers back to 
the same or a different entity (the antecedent) in text. 
\emph{Anaphora resolution} is the task to determine the antecedent
for a given anaphor. While \emph{direct anaphora resolution} attracts 
a lot of attention in the NLP community recently, 
such as Winograd Schema
Challenge \cite{rahman12-winograd,opitz-frank-2018,kocijan2019}, 
\emph{indirect anaphora resolution} or \emph{bridging anaphora resolution} 
is less well studied. 

In this paper, we focus on \emph{bridging anaphora resolution} where bridging anaphors and their antecedents are linked via various lexico-semantic, frame or encyclopedic relations. Following \newcite{houyufang13a} and \newcite{roesiger18a}, we mainly  consider 
``referential bridging'' in which bridging anaphors are truly anaphoric and bridging relations are context-dependent. In Example \ref{ex:bridging1}\footnote{All examples, if not specified
  otherwise, are from ISNotes \cite{markert12}.  Bridging
  anaphors are typed in boldface, antecedents in italics
  throughout this paper.}, both ``\underline{her building}'' 
and ``\emph{buildings with substantial damage}'' are plausible antecedent candidates for the bridging anaphor ``\textbf{residents}'' based on lexical semantics. In order to find the 
antecedent (\emph{buildings with substantial damage}), we have to take the meaning of  the broader discourse context into account.

\begin{examples}
\item \label{ex:bridging1} In post-earthquake parlance, \underline{her building} is a ``red''. 
After being inspected, \emph{buildings with substantial damage} were color-coded. Green allowed \textbf{residents} to re-enter; yellow allowed \textbf{limited access}; red allowed \textbf{residents} \textbf{one last entry} to gather everything they could within 15 minutes.
\end{examples}

Most previous studies on bridging anaphora resolution \cite{poesio04d,lassalle11,houyufang13a, houyufang18b} tackle the problem using the pairwise model and assume that the gold mention information is given. 
Most work \cite{poesio04d,lassalle11,houyufang13a} uses syntactic patterns to measure 
semantic relatedness between the head nouns of an anaphor and its antecedent. \newcite{houyufang18b} proposes a simple deterministic algorithm that also considers 
the semantics of modifications for head nouns. These approaches, however, do not take the broader context outside of noun phrases (i.e., anaphors and antecedent candidates) 
into account and often fail to resolve context-dependent bridging anaphors as demonstrated in Example \ref{ex:bridging1}.

Resolving bridging anaphors requires context-dependent text understanding.
Recently, \newcite{gardner2019qa} argue that question answering (QA) is a natural format to model tasks that require question understanding. 
In this paper, we cast bridging anaphora resolution 
as question answering based on context. We develop a 
QA system (\emph{BARQA}) for the task based on BERT \cite{devlin2018bert}. Given a context as shown in Example \ref{ex:bridging1}, 
we first rephrase every anaphor as a question, such as ``\emph{\textbf{residents} of what?}''. By answering the question, the system then identifies the span of the antecedent from the context. Compared to the pairwise model, our QA system does not require the gold or system mention information as the antecedent candidates. In addition, this framework allows us to integrate context 
outside of NPs when choosing antecedents for bridging anaphors. 
For instance, ``\underline{Green}'' and ``\underline{damage were color-coded}'' are among the top predicted answers for the above question.


Different from coreference resolution, there are no large-scale corpora available for 
referential bridging resolution due to its complexity. In this paper we propose a new method 
to generate a large amount of “quasi-bridging” training data from the automatically parsed 
Gigaword corpus \cite{gigaword5.0data,napoles12}. 
 We demonstrate that our “quasi-bridging” training data is a better pre-training choice 
for bridging anaphora resolution compared to the SQuAD corpus \cite{Rajpurkar16}. 
Moreover,  we show that our model pre-trained on this dataset and fine-tuned on a small amount of in-domain dataset achieves new state-of-the-art results for bridging anaphora resolution on two bridging corpora (i.e., ISNotes \cite{markert12} and BASHI \cite{roesiger18b}).

To summarize, the main contributions of our work are: (1) we formalize bridging anaphora 
resolution as 
a question answering problem and propose a QA model to solve the task; 
(2)  we explore a new method to generate a large amount of “quasi-bridging” training dataset and demonstrate its value for bridging anaphora resolution; and (3) we carefully carry out a series of experiments on two referential bridging corpora and 
provide some error analysis to verify the effectiveness of our QA model to resolve 
the context-dependent bridging anaphors in ISNotes. 
We release the code and all experimental datasets at \url{https://github.com/IBM/bridging-resolution}.

\section{Related Work}
\paragraph{Bridging Anaphora Resolution.}
Since the '90s, the empirical corpus studies related to bridging have been carried out on various genres and different languages 
\cite{fraurud90, poesio98, poesio04a, nissim04,gardent05,nedoluzhko09, eckart12, markert12, roesiger18b, poesio18}. 
Among those datasets, 
ISNotes \cite{markert12}, BASHI \cite{roesiger18b} and ARRAU \cite{poesio18} are recent three public English 
corpora which contain medium- to large-sized bridging annotations and have been used to evaluate 
systems' performance on bridging anaphora recognition \cite{houyufang13b, houyufang16, roesiger18a}, bridging anaphora resolution \cite{poesio04d, lassalle11, houyufang13a, houyufang18b}, as well as full bridging resolution \cite{houyufang14, houyufang18c, roesiger18a}. In this paper, we focus exclusively on the 
task of antecedent selection. 

\begin{figure*}[t]
\begin{center}
\includegraphics[width=0.99\textwidth]{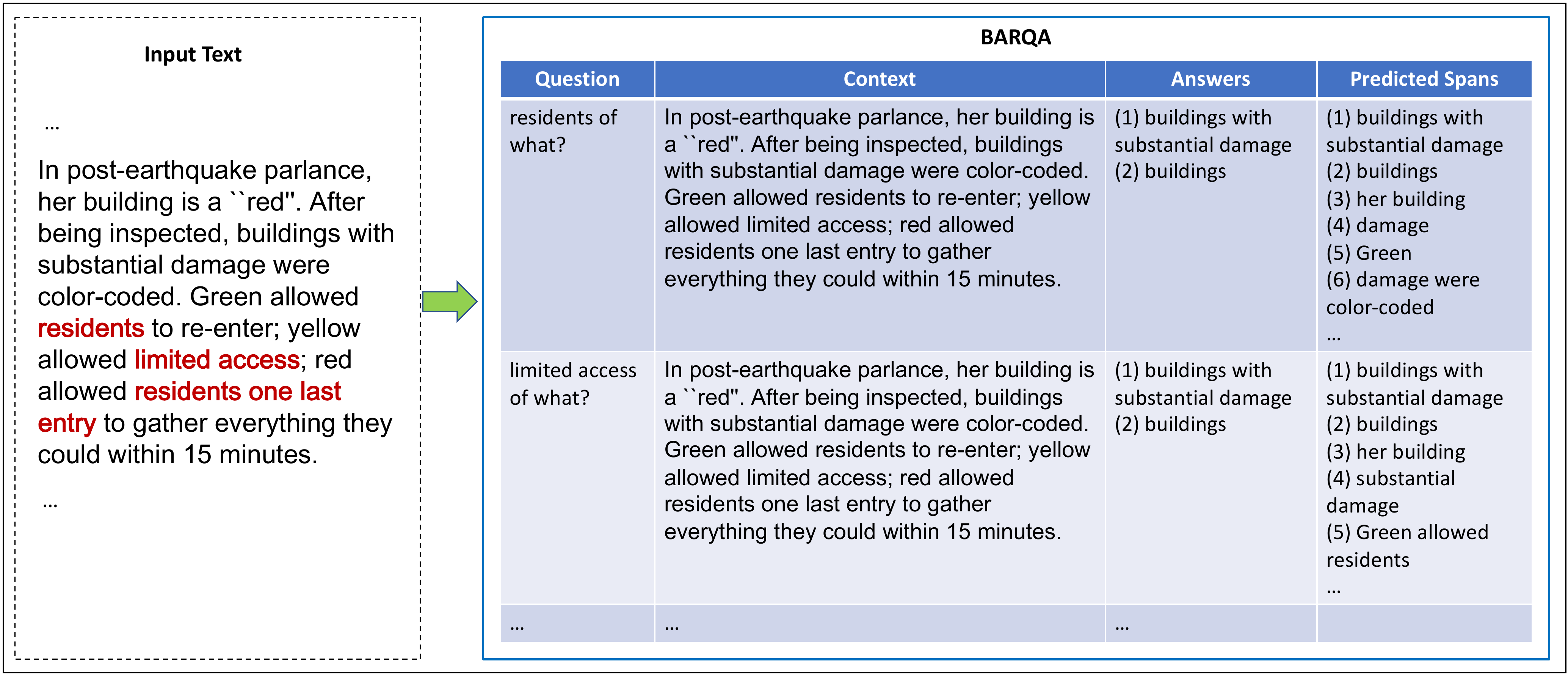}
\end{center}
\caption{Resolving bridging anaphors in Example \ref{ex:bridging1} using \emph{BARQA}.}
\label{fig:model}
\end{figure*}

It is worth noting that the bridging definition in the ARRAU corpus is different from the one used in the other two datasets. 
\newcite{roesiger18a} pointed out that 
ISNotes and BASHI contain ``referential bridging'' where bridging anaphors are truly anaphoric and bridging relations 
are context-dependent, while in ARRAU, most bridging links are purely lexical bridging pairs which are not context-dependent (e.g., 
\emph{Europe} -- \textbf{Spain} or \emph{Tokyo} -- \textbf{Japan}). 
In this paper, we focus on resolving referential bridging anaphors. 

Regarding the algorithm for bridging anaphora resolution, most previous work uses the pairwise model for the task. 
The model assumes gold or system mention information (NPs) is given beforehand. It creates (positive/negative) training instances  
by pairing every anaphor $a$ with its preceding mention $m$. Usually, $m$ is from a set of antecedent candidates which is formed using 
a fixed window size. 
\newcite{poesio04d} and \newcite{lassalle11} trained such pairwise models to resolve mereological bridging anaphors 
in the English GNOME corpus and the French DEDE corpus \cite{gardent05}, respectively. 
One exception is \newcite{houyufang13a}, which proposed a joint inference framework to resolve bridging anaphors in ISNotes. 
The framework
is built upon the pairwise model and predicts all semantically related bridging anaphors in one document together. 

Recently, \newcite{houyufang18b} generated a word representation resource 
 for bridging (i.e., \emph{embeddings\_bridging}) and proposed a simple 
 deterministic algorithm to find antecedents for bridging anaphors in ISNotes and BASHI. 
 The word representation resource is learned from a large corpus and it captures the common-sense knowledge (i.e., semantic relatedness) between NPs. 
 
 Different from the algorithms mentioned above, our QA model 
does not require the extracted or gold mentions (NPs) as the input, and it predicts 
the span of the antecedent for a bridging anaphor directly.

\paragraph{Question Answering.} \emph{Reading comprehension} or \emph{question answering based on context} has attacted much attention within the NLP community, in particular since \newcite{Rajpurkar16} 
released a large-scale dataset (SQuAD) consisting of 100,000+ questions on a set of paragraphs extracted from Wikipedia articles. 
Previous work has cast a few traditional NLP tasks as question answering, such as textual entailment \cite{McCann2018decaNLP}, entity--relation extraction \cite{li-etal-2019-entity}, and coreference resolution \cite{wu-acl2020}. However, unlike these tasks, we do not have large scale training datasets for bridging. As a result, we form the questions for our task in a more natural way in order to leverage the existing 
QA datasets (e.g., SQuAD) that require common-sense reasoning. In addition, we generate a large-scale training dataset of “quasi-bridging” and demonstrate that it is a good pre-training corpus for bridging anaphora resolution. 

Recently, \newcite{gardner2019qa} argue that we should consider question answering as 
a \emph{format} instead of a \emph{task} in itself. From this perspective, our work can be seen as a specific probing task to test a QA model's ability to understand bridging anaphora based on context.


\paragraph{Winograd Schema Challenge.} Bridging anaphora resolution shares   
some similarities with Winograd Schema Challenge (WSC). Specifically, in both tasks, one has to understand the context to find the antecedents for anaphors. 
However,  the antecedent search space in bridging anaphora resolution is much bigger 
than the one in WSC. 
This is because an anaphor (pronoun) and its antecedent in 
WSC are usually from the same sentence, while 
bridging pairs usually require cross-sentence inference. 
For instance, in ISNotes, only around 26\% of anaphors have antecedents 
occurring in the same sentence, and 23\% of anaphors have antecedents that are more than two sentences away. 

Recently, \newcite{kocijan2019} use some heuristics to generate a large-scale WSC-like dataset and report that the model pre-trained on this dataset achieves the best results on several WSC datasets after being fine-tuned on a small 
in-domain dataset. We find similar patterns of results
for bridging anaphora resolution (see Section \ref{sec:result1_lenient}).

\section{BARQA: A QA System for Bridging Anaphora Resolution}
\label{sec:model}
In this section, we describe our QA system (called \emph{BARQA}) for bridging anaphora resolution 
in detail. Figure \ref{fig:model} illustrates how \emph{BARQA} predicts antecedents for bridging anaphors in Example \ref{ex:bridging1}.


\subsection{Problem Definition}
We formulate bridging anaphora resolution as a context-based QA problem. 
More specifically, 
given a bridging anaphor $a$ and its surrounding context $c_a$, 
we rephrase $a$ as a question $q_a$. 
The goal is to predict a text span $s_a$ from $c_a$ that is the antecedent of $a$. 
We propose to use the span-based QA framework 
to extract $s_a$. 
In general, our \emph{BARQA} system is built on top of the vanilla BERT QA framework \cite{devlin2018bert}. We further modify the inference algorithm to guarantee that the answer 
span $s_a$ should always appear before the bridging anaphor $a$ (see Section \ref{sec:infer} for more details).

Following \newcite{devlin2018bert}, we present the input question $q_a$ and the context $c_a$ as a single packed sequence 
``\emph{$[cls] \: q_a \: [sep] \: c_a$}'' and calculate the probabilities of every word 
in $c_a$ being the start and end of the answer span. The training objective is the log-likelihood of the correct start and end positions.

\subsection{Question Generation}
In English, the preposition ``\emph{\textbf{of}}'' in the syntactic structure 
``\emph{$np_1$ \textbf{of} $np_2$}'' encodes different associative relations 
between noun phrases that cover a variety of bridging relations. For instance, ``\emph{the chairman of IBM} '' indicates \emph{a professional function in an organization}, and ``\emph{the price of the stock}'' indicates \emph{an attribute of an object}. \newcite{poesio04d} also used such patterns
to estimate the part-of bridging relations. 
These patterns 
reflect how we explain bridging anaphora as human beings.
It seems that the most natural way to understand 
the meaning of a bridging anaphor $a$ 
is to find the answer for the question ``\emph{$a$ of what?}'' from the surrounding context of $a$. 

As a result, in order to generate the corresponding question $q_a$ for 
a bridging anaphor $a$, we first create $a'$ by removing all words appearing after the head of $a$, we then concatenate $a'$ with 
``\emph{of what?}'' to form the question. This is because, as 
pointed by \newcite{houyufang18b}, premodifiers of bridging anaphors are essential elements to understand bridging relations.
For instance, for the 
bridging anaphor ``\textbf{a painstakingly documented report, based on hundreds of interviews with randomly selected refugees}'',
the corresponding question is ``\emph{a painstakingly documented report of what?}''.

\begin{figure*}[t]
\begin{center}
\includegraphics[width=0.99\textwidth]{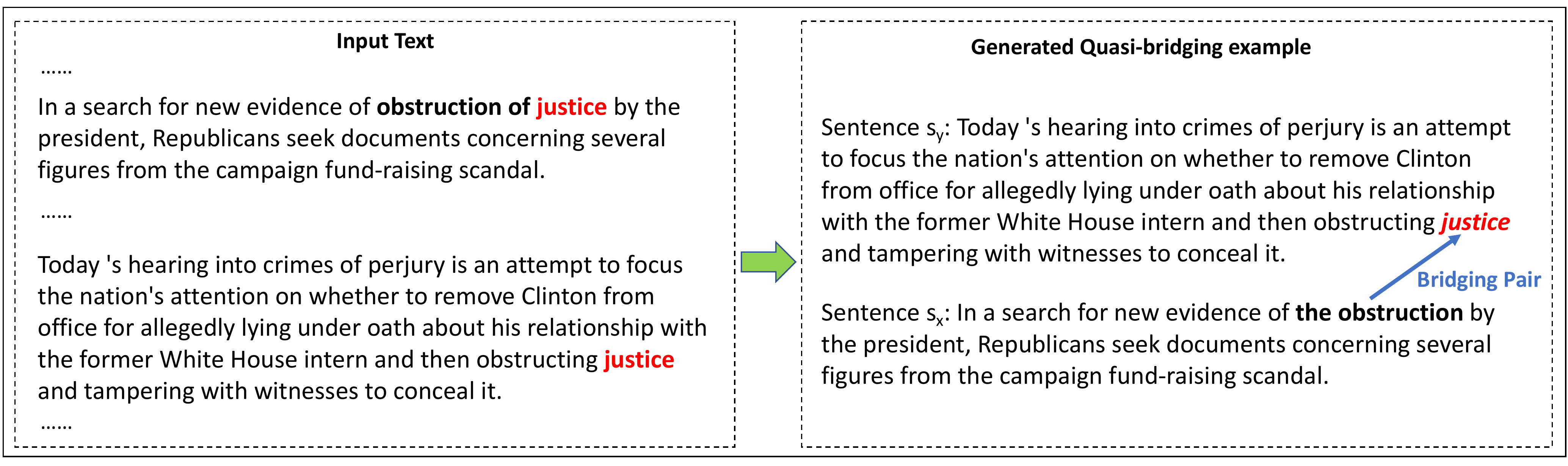}
\end{center}
\caption{Examples of generating ``quasi-bridging'' training data.}
\label{fig:trainingdata}
\end{figure*}


\subsection{Answer Generation}
\label{sec:answer}
For each bridging anaphor $a$ together with its corresponding question $q_a$ and context $c_a$ described above, we construct a list of answers $A$
that contains all antecedents of $a$ occurring in the context $c_a$.\footnote{In ISNotes 
and BASHI, we use gold coreference annotations from OntoNotes \cite{ontonotes4.0data} to identify all possible antecedents for every bridging anaphor.}
In addition, for every NP antecedent $n$ from $A$, we add the following variations which represent the main semantics of $n$ into the answer list:
\begin{itemize}
\item the head of $n$ (e.g., \emph{last week's \underline{earthquake}})
\item $n'$ which is created by removing all postmodifiers from $n$ (e.g., \emph{\underline{the preliminary conclusion} 
from a survey of 200 downtown high-rises}) 
\item $n''$ which is created by removing all postmodifiers and the determiner from $n$ (e.g., \emph{the \underline{total potential claims} from the disaster})
\end{itemize}

It is worth noting that if the context $c_a$ does not contain any antecedent 
for the bridging anaphor $a$ (e.g., some anaphors do not have antecedents occurring in $c_a$ if 
we use a small window size to construct it), we put ``\emph{no answer}'' into the answer list $A$.

\subsection{Inference}
\label{sec:infer}
Different from the SQuAD-style question answering where 
there is no specific requirement for the position of the predicted span,  
in bridging anaphora resolution, an anaphor must appear after its antecedent. 
Therefore in the inference stage, for each bridging anaphor $a$, we first identify the position of $a$ in its context $c_a$, then we only predict text spans 
which appear before $a$. We further prune the list of predicted text spans by only keeping the top $k$ span candidates that contain at most $l$ words ($k$ and $l$ are empirically set to 20 and 5, respectively). We also prune span predictions that are function words (e.g., \emph{a, an, the, this, that}).

\subsection{Training}
 During the training process, we first use SpanBERT \cite{m2019spanbert} to initialize our BARQA model because it shows promising improvements on SQuAD 1.1 compared to the vanilla BERT embeddings. We then continue to train our model using different pre-training and fine-tuning strategies. Section \ref{sec:result1_lenient} describes different training strategies in detail.
  
For every training strategy, we train BARQA for five epochs with a learning rate of 
3e-5 and a batch size of 24.\footnote{In general, the small learning rate (i.e., 3e-5, 4e-5, and 5e-5) and small fine-tuning epochs are common practices for fine-tuning BERT models. We test the combination of these parameters for various training configurations on a small set (10 documents) of the ISNotes corpus and the BASHI corpus, 
respectively. On both corpora, we observed that a learning rate of 3e-5, 4e-5, or 5e-5 has minimal impact on results; and for each learning rate, the result continues improving
at the beginning (epochs = 1,2,3,4,5), but the performances stays more or less the same after epochs$>$5.} During training and testing, the maximum
text length is set to 128 tokens.

\section{Generate ``Quasi-bridging'' Training Data}
\label{sec:trainingdata}
Bridging anaphora is a complex phenomenon, and there are no large-scale 
corpora available for referential bridging. In this section, we describe how we 
generate a large scale ``quasi-bridging'' dataset. 

\newcite{houyufang18} explores the syntactic prepositional and possessive structures of NPs to train word embeddings for bridging. Inspired by this work,
we first use these structures to identify	``bridging anaphors'' and the corresponding ``antecedents''. Next, we map them back to the discourse to create bridging-like examples.


More specifically, given a text, we first extract NPs 
containing the prepositional structure (e.g.,  \textbf{X} \emph{preposition} \textbf{Y}) or the possessive structure (e.g., \textbf{Y} \emph{'s} \textbf{X}).
In order to have a high-quality set of automatically generated bridging annotations, we apply an additional constraint to the above NPs, i.e.,  
\textbf{X} and \textbf{Y} should not contain any other NP nodes in the constituent tree. For instance, we do not consider 
NPs such as ``\emph{the political value of imposing sanctions against South Africa}'' or
``\emph{the cost of repairing the region's transportation system}''.

Figure \ref{fig:trainingdata} illustrates how we generate a bridging annotation 
with a sentence pair 
\{$s_y$, $s_x$\} from a raw text\footnote{The raw text is from the Gigaword corpus \cite{gigaword5.0data,napoles12}.}: we first extract the NP ``\emph{obstruction of justice}'' from the sentence $s_i$ and identify \textbf{X}/\textbf{Y} in this extracted NP (i.e., \textbf{X} = obstruction, \textbf{Y} = justice). Next, we collect a list of sentences $S$ from the whole text. Every sentence in $S$ contains \textbf{$\textbf{Y}$} but does not contain \textbf{$\textbf{X}$}. If $S$ contains more than one sentence, we choose the one 
which 
is the closest to $s_i$ as $s_y$. 
This is because close sentences are more likely semantically related. Finally, 
we generate the sentence $s_x$ by replacing ``\emph{obstruction of justice}'' in the original sentence $s_i$ with ``\emph{\textbf{the obstruction}}''.
This gives us a quasi-bridging example with two adjacent sentences (i.e., $s_y$ and $s_x$) and a bridging link (i.e., \emph{justice} - \textbf{the obstruction}).

As a result, we obtain a large amount of ``quasi-bridging'' training data (i.e., around 2.8 million bridging pairs) by applying the method described above to the NYT19 section of the automatically 
parsed Gigaword corpus.

In order to understand the quality of our ``quasi-bridging'' training dataset, we randomly sample 
100 quasi-bridging sentence pairs and manually check bridging annotations in these instances. We score each bridging annotation using a scale of 0-2: ``2'' means that the bridging annotation is correct and the sentence pair sounds natural; ``1'' indicates that the example makes sense, but it does not sound natural in English; and ``0'' denotes that the annotation is unacceptable. Overall, 
we find that 25\% of instances and 37\% of instances have a score of 2 and 1, respectively.
 And the remaining 38\% of instances are scored as zero. In general, our noisy ``quasi-bridging'' training dataset does contain a large number of  diverse bridging pairs.




\section{Experiments}
\label{sec:exp}
\subsection{Datasets}
\label{sec:ds}
We use four datasets for experiments. The first dataset is ISNotes\footnote{\url{http://www.h-its.org/en/research/nlp/isnotes-corpus}} released by 
\newcite{markert12}. This dataset contains 50 texts with 663 referential bridging NPs from the World Street Journal (WSJ) portion of the OntoNotes corpus \cite{ontonotes4.0data}.
The second dataset is called BASHI from \newcite{roesiger18b}. It contains 
459 bridging NPs\footnote{BASHI considers comparative anaphora as bridging anaphora. We exclude them from this study.} with 344 referential anaphors from 50 WSJ texts\footnote{Note that these WSJ articles are different from the ones in ISNotes.}. Note that bridging
anaphors in these two corpora are not limited to definite NPs as in previous work \cite{poesio97b,lassalle11} and bridging relations are not limited to the prototypical \emph{whole -- part} relation or \emph{set -- element} 
relation. We consider these two corpora as expert-annotated in-domain 
datasets.

We assume that some reasoning skills (e.g., world knowledge, word relatedness) required to answer questions in SQuAD 
can also be applied for bridging anaphora resolution. Therefore we include 
the SQuAD 1.1 training data \cite{Rajpurkar16} as one training dataset. 
Another training dataset is the large scale quasi-bridging corpus (\emph{QuasiBridging}) described in Section \ref{sec:trainingdata}.

Table \ref{tab:datasets} summarizes the four datasets mentioned above. 
Note that in ISNotes and BASHI, the number of QA pairs is more than the number of bridging anaphors. This is because an anaphor can have multiple 
antecedents (e.g., coreferent mentions of the same antecedent entity).

\begin{table*}[th]
\begin{center}
\begin{tabular}{l|l|c|c|c}
\hline
\textbf{Corpus}&\textbf{Genre}&\textbf{Bridging Type} &\textbf{\# of Anaphors}&\textbf{\# QA Pairs}\\ 
\hline
\emph{ISNotes}&WSJ news articles&referential bridging&663&1,115\\
\emph{BASHI}&WSJ news articles&referential bridging&344&486\\
\emph{SQuAD 1.1 (train)}&Wikipedia paragraphs& - &-&87,599\\
\emph{QuasiBridging}&NYT news articles&quasi bridging &2,870,274&2,870,274\\
\hline
\end{tabular}
\end{center}
\caption{\label{tab:datasets} Four datasets used for experiments.}
\end{table*}


\subsection{Experimental Setup}
\label{sec:setup}
Following \newcite{houyufang18b}, we use \emph{accuracy}\ on the number of bridging 
anaphors to measure systems' performance for resolving bridging anaphors on ISNotes and BASHI. It is calculated as the number of the correctly resolved bridging anaphors divided by the total number of bridging anaphors.

We measure two types of \emph{accuracy}: \emph{lenient accuracy} and \emph{strict accuracy}. In \emph{strict accuracy}, only the original gold antecedent annotations are counted as the correct answers. 
For \emph{lenient accuracy}, we add the additional variations of the original 
antecedent annotations (described in Section \ref{sec:answer}) into the correct answer list. For instance, suppose that the gold antecedent annotation is
``\emph{the Four Seasons restaurant}'', and the predicted span is 
``\emph{Four Seasons restaurant}'', we count this prediction as an incorrect 
prediction in \emph{strict accuracy} evaluation. However, it is a correct prediction in \emph{lenient accuracy} evaluation. 

It is worth noting that our lenient accuracy corresponds to the ``exact match'' metric in SQuAD \cite{Rajpurkar16}. The correct answer lists that are generated as described in Section \ref{sec:answer} can partially address the evaluation problem of imperfect system mention predictions. We do not report F1 score because it will give 
partial credit for a prediction that does not capture the main semantics of the original gold annotation, such as ``\emph{the Four Seasons}''.

During evaluation, for every bridging anaphor $a$, let $s_a$ be the sentence containing $a$, 
we use the first sentence of the text, 
the previous two sentences of $s_a$, as well as $s_a$ 
to form $a$'s surrounding context $c_a$. 
This is in line with \newcite{houyufang18b}'s antecedent candidate selection strategy. 

\begin{table*}[t]
\begin{center}
\begin{tabular}{l|c|c}
\hline
\textbf{BARQA}& \textbf{Lenient Accuracy on ISNotes} & \textbf{Lenient Accuracy on BASHI}\\ 
\hline
\multicolumn{3}{c}{\textbf{Large-scale (out-of-domain/noisy) training data}}\\ \hline
\emph{SQuAD 1.1}&28.81&29.94\\ \hline
\emph{QuasiBridging}&25.94&17.44\\
\hline
\multicolumn{3}{c}{\textbf{Small in-domain training data}}\\ \hline
\emph{BASHI}&38.16& - \\ \hline
\emph{ISNotes}&-& 35.76 \\ \hline
\multicolumn{3}{c}{\textbf{Pre-training + In-domain fine-tuning}}\\ \hline
\emph{SQuAD 1.1 + BASHI}&42.08&-\\ \hline
\emph{QuasiBridging + BASHI}&\textbf{ 47.21}$\star$&-\\ \hline
\emph{SQuAD 1.1 + ISNotes}&-&35.76\\ \hline
\emph{QuasiBridging + ISNotes}&-&\textbf{37.79}\\ 
\hline
\end{tabular}
\end{center}
\caption{\label{tab:qa_result1} Results of \emph{BARQA} on ISNotes 
and BASHI using different training strategies. $\star$ indicates
statistically significant differences over the other models (two-sided paired approximate randomization test, $p < 0.01$).}
\end{table*}

\subsection{Results on ISNotes and BASHI Using Different Training Strategies}
\label{sec:result1_lenient}
In this section, we carry out experiments using our \emph{BARQA} system with different training strategies. For every bridging anaphor $a$, we choose the span with the highest confidence score from its context $c_a$ as the answer for the question $q_a$ and use this span as the predicted antecedent.
 We report results on ISNotes and BASHI using 
\emph{lenient accuracy} (see Table \ref{tab:qa_result1}).

Looking at the results on ISNotes, we find that \emph{BARQA} 
 trained on a small number of 
in-domain dataset (\emph{BASHI}) achieves an accuracy of 38.16\% on ISNotes, 
which is better than the model trained on the other two large-scale datasets (\emph{SQuAD 1.1} and \emph{QuasiBridging}). However, when using these two datasets to pre-train the model then fine-tuning it with the small in-domain 
dataset (\emph{BASHI}), both settings (i.e., \emph{SQuAD 1.1 + BASHI} and 
\emph{QuasiBridging + BASHI}) achieve better results compared to using \emph{BASHI} as the only training dataset. 
This verifies the value of the \emph{pre-training + fine-tuning} strategy, i.e.,
pre-training the model with large scale out-of-domain or noisy dataset, then fine-tuning it with a small in-domain dataset.

Particularly, we notice that the performance of using \emph{QuasiBridging} alone is worse than the one using \emph{SQuAD 1.1} only. However,  combining \emph{QuasiBridging} and \emph{BASHI} achieves the best result
on ISNotes, with an accuracy of 47.21\%. It seems that the large-scale in-domain noisy training data (\emph{QuasiBridging}) brings more value than the large-scale out-of-domain training data (\emph{SQuAD 1.1}). 

We observe similar patterns on the results on BASHI. 
Pre-training the model on \emph{QuasiBridging} then fine-tuning it on \emph{ISNotes} achieves the best result with an accuracy of 37.79\%. 
Furthermore, when evaluating on BASHI, it seems that using \emph{SQuAD 1.1} as the pre-training dataset does not bring additional values
when combining it with \emph{ISNotes}.

\begin{table*}[thb]
\begin{center}
\begin{tabular}{l|c|c}
\textbf{System} &\textbf{Use Gold Mentions}&\textbf{Accuracy}\\ 
\hline
\multicolumn{3}{c}{\textbf{Models from \newcite{houyufang13a}}}\\ \hline
\emph{pairwise model III}&yes&36.35\\
\emph{MLN model II}&yes&41.32\\
\hline
\multicolumn{3}{c}{\textbf{Models from \newcite{houyufang18b}}}\\ \hline
\emph{embeddings\_bridging (NP head + modifiers)}&yes&39.52\\
\emph{MLN model II + embeddings\_bridging (NP head + modifiers)}&yes&46.46\\ \hline
\multicolumn{3}{c}{\textbf{This work}}\\ \hline
\emph{BARQA with gold mentions/semantics, strict accuracy}&yes&\textbf{50.08}\\
\emph{BARQA without mention information, strict accuracy}&no&36.05\\ 
\emph{BARQA without mention information, lenient accuracy}&no&47.21\\ \hline

\end{tabular}
\end{center}
\caption{\label{tab:qa_result2} Results of different systems for bridging anaphora resolution in ISNotes. Bold indicates
statistically significant differences over the other models (two-sided paired approximate randomization test, $p < 0.01$).}
\end{table*}

\begin{table*}[thb]
\begin{center}
\begin{tabular}{l|c|c}
\textbf{System} &\textbf{Use System Mentions}&\textbf{Accuracy}\\ 
\hline
\multicolumn{3}{c}{\textbf{Model from \newcite{houyufang18b}}}\\ \hline
\emph{embeddings\_bridging (NP head + modifiers)}&yes&29.94\\ \hline
\multicolumn{3}{c}{\textbf{This work}}\\ \hline

\emph{BARQA with system mentions/semantics, strict accuracy}&yes&\textbf{38.66}\\
\emph{BARQA without mention information, strict accuracy}&no&32.27\\ 
\emph{BARQA without mention information, lenient accuracy}&no&\textbf{37.79}\\ \hline

\end{tabular}
\end{center}
\caption{\label{tab:qa_result3} Results of different systems for bridging anaphora resolution in BASHI. Bold indicates
statistically significant differences over the other models (two-sided paired approximate randomization test, $p < 0.01$).}
\end{table*}

\subsection{Results on ISNotes and BASHI Compared to Previous Approaches}
Previous work for bridging anaphora resolution on ISNotes and BASHI
use gold/system mentions as antecedent candidates and report results using \emph{strict accuracy} \cite{houyufang13a, houyufang18b}. 

In order to fairly compare against these systems, for every bridging anaphor $a$, we first map all top 20 span predictions 
of our system \emph{BARQA} to the gold/system mentions, then we choose the gold/system mention with the highest confidence score as the predicted antecedent. Specifically, we map a predicted span $s$ to a mention $m$ if they share the same head 
and $s$ is part of $m'$ ($m'$ is created by removing all postmodifiers from $m$). For instance, ``\emph{total potential claims}'' is mapped to the
mention ``\emph{the total potential claims from the disaster}''.
If a predicted span can not be mapped to any gold/system mentions, we filter it out.  Following \newcite{houyufang18b}, 
we only keep the predictions whose semantic types are ``time'' if $a$ is a time expression. 
The above process 
is equal to using gold/system mentions and their semantic information to further prune \emph{BARQA}'s span predictions.

Table \ref{tab:qa_result2} and Table \ref{tab:qa_result3} compare the results 
of our system \emph{BARQA} against previous studies for   
bridging anaphora resolution on ISNotes and BASHI, respectively. 
For both datasets, the \emph{BARQA} model is trained 
using the best strategy reported in Table 
\ref{tab:qa_result1} (pre-training with \emph{QuasiBridging} + fine-tuning with small in-domain data).

On ISNotes, previously \newcite{houyufang18b} reported the best result by adding the prediction from a deterministic algorithm 
(\emph{embeddings\_bridging (NP head + modifiers)}) as an additional feature into the global inference model (\emph{MLN II}) proposed by \newcite{houyufang13a}. The 
deterministic algorithm is based on word embeddings for bridging and models the meaning of an NP based on its head noun and modifications. 

Our system \emph{BARQA}, when using the gold mentions together with their 
semantic information to further prune the span predictions, achieves the new state-of-the-art result on ISNotes, with a strict accuracy of 50.08\% (see \emph{BARQA with gold mentions/semantics, strict accuracy} in Table \ref{tab:qa_result2}). However, we argue that using gold mention information to construct the set of antecedent candidates is a controlled experiment condition,
and our experiment setup \emph{BARQA without mention information, lenient accuracy} is a more realistic scenario in practice. 

On BASHI, \newcite{houyufang18b} reported an accuracy of 29.94\% (\emph{strict accuracy}) using automatically extracted mentions from the 
gold syntactic tree annotations. Our system \emph{BARQA} without any mention/semantic information achieves an accuracy of 32.27\% using 
the same \emph{strict accuracy} evaluation. The result of \emph{BARQA} is 
further improved with an accuracy of 38.66\% when we integrate mention/semantic information into the model.

Note that \newcite{houyufang18b} also adapted their deterministic algorithm 
to resolve lexical bridging anaphors on ARRAU \cite{poesio18} and reported an 
accuracy of 32.39\% on the \emph{RST Test} dataset. Although in this paper we do not focus on lexical bridging, our model \emph{BARQA} can also be applied to 
resolve lexical bridging anaphors. We found that \emph{BARQA} trained on the 
\emph{RST Train} dataset alone with around 2,000 QA pairs achieves an accuracy of 34.59\% on the \emph{RST Test} dataset.

\section{Error Analysis} 
In order to better understand our model, 
we automatically label bridging anaphors 
in ISNotes as 
either ``\emph{referential bridging/world-knowledge}'' or ``\emph{referential bridging/context-dependent}''. We then analyze the performance of \emph{BARQA}
and the best model from \newcite{houyufang18b} on these two categories.

\newcite{roesiger18a} pointed out that although lexical and referential bridging are two different  concepts, sometimes they can co-occur within the same pair of expressions. In Example \ref{ex:bridging2}, ``\textbf{Employees}'' is an 
anaphoric expression. At the same time, the relation between the antecedent entity ``\{\emph{Mobil Corp./the company's}\}'' and the bridging anaphor 
``\textbf{Employees}'' corresponds to the common-sense world knowledge which is true without any specific context. We call such cases as ``\emph{referential bridging/world-knowledge}''.
Differently, we call a bridging anaphor as ``\emph{referential bridging/context-dependent}'' if it has multiple equally plausible antecedent candidates according to the common-sense world knowledge about the NP pairs and we have to analyze the context to choose the antecedent (see Example {\ref{ex:bridging1}}). One may argue that ``\{the exploration and production division -- \textbf{Employees}\}'' in Example \ref{ex:bridging2} is also a valid common-sense knowledge fact, however, we consider that it is less prominent than ``\{\emph{the company's} -- \textbf{Employees}\}''.

\begin{examples}
\item \label{ex:bridging2}  \emph{Mobil Corp.} is preparing to slash the size of its workforce in the U.S., possibly as soon as next month, say individuals familiar with \emph{the company's} strategy. The size of the cuts isn't known, 
but they'll be centered in the exploration and production division, which is responsible for locating oil reserves, drilling wells and pumping crude oil and natural gas. \textbf{Employees} haven't yet been notified.
\end{examples}

For a bridging anaphor $a$, the deterministic algorithm (\emph{embeddings\_bridging}) from \newcite{houyufang18b} uses a word representation resource learned from a large corpus to predict the most semantically related NP among all NP candidates as the antecedent. The predictions from this system reflect 
the common-sense world knowledge about the NP pairs. We thus use this algorithm to label bridging anaphors in ISNotes: if a bridging anaphor is correctly resolved by \emph{embeddings\_bridging}, we label it as ``\emph{referential bridging/world-knowledge}'', otherwise the label is 
``\emph{referential bridging/context-dependent}''.


Table \ref{tab:qa_result4} compares the percentage of correctly resolved anaphors between \emph{BARQA} with gold mentions and the best model from \newcite{houyufang18b} (\emph{MLN II + emb}) on the two bridging categories. Note that \emph{MLN II + emb} 
contains several context-level features 
(e.g., document span, verb pattern). 
Overall, it seems that our \emph{BARQA} model is better at resolving 
context-dependent bridging anaphors.



\begin{table}[t]
\begin{center}
\begin{tabular}{l|l|c|c}
 &\# pairs&\emph{BARQA}&\emph{MLN II + emb}\\  \hline
\textbf{Know.}& 256&71.88&\textbf{88.28}\\ \hline
\textbf{Context}&407&\textbf{36.36}&19.90\\ \hline
\end{tabular}
\end{center}
\caption{\label{tab:qa_result4} Comparison of the percentage of correctly resolved anaphors between \emph{BARQA} and the best model from \newcite{houyufang18b}
  on two bridging categories. }
\end{table}

\section{Conclusions}
In this paper, we model bridging anaphora resolution as a question answering problem and propose a QA system (\emph{BARQA}) to solve the task. 

We also propose a new method to automatically generate a large scale of ``quasi-bridging''  training data. We show that our QA system, when trained on this ``quasi-bridging'' training dataset and fine-tuned on a small amount of in-domain dataset, achieves the new state-of-the-art results on two bridging corpora.

Compared to previous systems, our model is simple and more realistic in practice: it does not require any gold annotations to construct the list of antecedent candidates. Moreover, under the proposed QA formulation, 
our model can be easily strengthened by adding other span-based text understanding QA corpora as pre-training datasets. 

Finally, we will release our experimental QA datasets (in the SQuAD json format) for bridging anaphora resolution on ISNotes and BASHI. They can be used to test a QA model's ability to understand a text in terms of bridging inference.








\section*{Acknowledgments}
The author appreciates the valuable feedback from the anonymous reviewers.

\bibliographystyle{acl_natbib}
\bibliography{../../bib/lit/lit}

\begin{thebibliography}{34}
\expandafter\ifx\csname natexlab\endcsname\relax\def\natexlab#1{#1}\fi

\bibitem[{Devlin et~al.(2019)Devlin, Chang, Lee, and
  Toutanova}]{devlin2018bert}
Jacob Devlin, Ming{-}Wei Chang, Kenton Lee, and Kristina Toutanova. 2019.
\newblock {BERT:} pre-training of deep bidirectional transformers for language
  understanding.
\newblock In \emph{Proceedings of the 2019 Conference of the North American
  Chapter of the Association for Computational Linguistics: Human Language
  Technologies, {\em Minneapolis, USA, 2--7 June 2019}}, pages 4171--4186.

\bibitem[{Eckart et~al.(2012)Eckart, Riester, and Schweitzer}]{eckart12}
Kerstin Eckart, Arndt Riester, and Katrin Schweitzer. 2012.
\newblock A discourse information radio news database for linguistic analysis.
\newblock In Christian Chiarcos, Sebastian Nordhoff, and Sebastian Hellmann,
  editors, \emph{Linked Data in Linguistics}, pages 65--76. Springer Berlin
  Heidelberg.

\bibitem[{Fraurud(1990)}]{fraurud90}
Kari Fraurud. 1990.
\newblock Definiteness and the processing of noun phrases in natural discourse.
\newblock \emph{Journal of Semantics}, 7:395--433.

\bibitem[{Gardent and Manu{\'e}lian(2005)}]{gardent05}
Claire Gardent and H{\'e}l{\`e}ne Manu{\'e}lian. 2005.
\newblock Cr{\'e}ation d'un corpus annot{\'e} pour le traitement des
  descriptions d{\'e}finies.
\newblock \emph{Traitement Automatique des Langues}, 46(1):115--140.

\bibitem[{Gardner et~al.(2019)Gardner, Berant, Hajishirzi, Talmor, and
  Sewon}]{gardner2019qa}
Matt Gardner, Jonathan Berant, Hannaneh Hajishirzi, Alon Talmor, and Min Sewon.
  2019.
\newblock Question answering is a format; when is it useful?
\newblock \emph{arXiv preprint arXiv:909.11291}.

\bibitem[{Hou(2016)}]{houyufang16}
Yufang Hou. 2016.
\newblock Incremental fine-grained information status classification using
  attention-based {LSTM}s.
\newblock In \emph{Proceedings of the 26th International Conference on
  Computational Linguistics, {\em Osaka, Japan, 11--16 December 2016}}, pages
  1880--1890.

\bibitem[{Hou(2018{\natexlab{a}})}]{houyufang18b}
Yufang Hou. 2018{\natexlab{a}}.
\newblock A deterministic algorithm for bridging anaphora resolution.
\newblock In \emph{Proceedings of the 2018 Conference on Empirical Methods in
  Natural Language Processing, {\em Brussels, Belgium, 31 October-- 4 November
  2018}}, pages 1938--1948.

\bibitem[{Hou(2018{\natexlab{b}})}]{houyufang18}
Yufang Hou. 2018{\natexlab{b}}.
\newblock \href {https://arxiv.org/pdf/1803.04790.pdf} {Enhanced word
  representations for bridging anaphora resolution}.
\newblock In \emph{Proceedings of the 2018 Conference of the North American
  Chapter of the Association for Computational Linguistics: Human Language
  Technologies, {\em New Orleans, Louisiana, 1--6 June 2018}}, pages 1--7.

\bibitem[{Hou et~al.(2013{\natexlab{a}})Hou, Markert, and
  Strube}]{houyufang13b}
Yufang Hou, Katja Markert, and Michael Strube. 2013{\natexlab{a}}.
\newblock \href {http://aclweb.org/anthology/D13-1077.pdf} {Cascading
  collective classification for bridging anaphora recognition using a rich
  linguistic feature set}.
\newblock In \emph{Proceedings of the 2013 Conference on Empirical Methods in
  Natural Language Processing, {\em Seattle, Wash., 18--21 October 2013}},
  pages 814--820.

\bibitem[{Hou et~al.(2013{\natexlab{b}})Hou, Markert, and
  Strube}]{houyufang13a}
Yufang Hou, Katja Markert, and Michael Strube. 2013{\natexlab{b}}.
\newblock \href {http://aclweb.org/anthology/N13-1111.pdf} {Global inference
  for bridging anaphora resolution}.
\newblock In \emph{Proceedings of the 2013 Conference of the North American
  Chapter of the Association for Computational Linguistics: Human Language
  Technologies, {\em Atlanta, Georgia, 9--14 June 2013}}, pages 907--917.

\bibitem[{Hou et~al.(2014)Hou, Markert, and Strube}]{houyufang14}
Yufang Hou, Katja Markert, and Michael Strube. 2014.
\newblock \href {http://aclweb.org/anthology/D13-1077.pdf} {A rule-based system
  for unrestricted bridging resolution: Recognizing bridging anaphora and
  finding links to antecedents}.
\newblock In \emph{Proceedings of the 2014 Conference on Empirical Methods in
  Natural Language Processing, {\em Doha, Qatar, 25--29 October 2014}}, pages
  2082--2093.

\bibitem[{Hou et~al.(2018)Hou, Markert, and Strube}]{houyufang18c}
Yufang Hou, Katja Markert, and Michael Strube. 2018.
\newblock Unrestricted bridging resolution.
\newblock \emph{Computational Linguistics}, 44(2):237--284.

\bibitem[{Joshi et~al.(2019)Joshi, Chen, Liu, Weld, Zettlemoyer, and
  Levy}]{m2019spanbert}
Mandar Joshi, Danqi Chen, Yinhan Liu, Daniel~S. Weld, Luke Zettlemoyer, and
  Omer Levy. 2019.
\newblock \href {http://arxiv.org/abs/1907.10529} {Spanbert: Improving
  pre-training by representing and predicting spans}.
\newblock \emph{arXiv preprint arXiv:1907.10529}.

\bibitem[{Kocijan et~al.(2019)Kocijan, Cretu, Camburu, Yordanov, and
  Lukasiewicz}]{kocijan2019}
Vid Kocijan, Ana-Maria Cretu, Oana-Maria Camburu, Yordan Yordanov, and Thomas
  Lukasiewicz. 2019.
\newblock A surprisingly robust trick for the {W}inograd {S}chema {C}hallenge.
\newblock In \emph{Proceedings of the 57th Annual Meeting of the Association
  for Computational Linguistics, {\em Florence, Italy, 28 July--2 August
  2019}}, pages 4837--4842.

\bibitem[{Lassalle and Denis(2011)}]{lassalle11}
Emmanuel Lassalle and Pascal Denis. 2011.
\newblock Leveraging different meronym discovery methods for bridging
  resolution in {F}rench.
\newblock In \emph{Proceedings of the 8th Discourse Anaphora and Anaphor
  Resolution Colloquium (DAARC 2011), {\em Faro, Algarve, Portugal, 6--7
  October 2011}}, pages 35--46.

\bibitem[{Li et~al.(2019)Li, Yin, Sun, Li, Yuan, Chai, Zhou, and
  Li}]{li-etal-2019-entity}
Xiaoya Li, Fan Yin, Zijun Sun, Xiayu Li, Arianna Yuan, Duo Chai, Mingxin Zhou,
  and Jiwei Li. 2019.
\newblock Entity-relation extraction as multi-turn question answering.
\newblock In \emph{Proceedings of the 57th Annual Meeting of the Association
  for Computational Linguistics, {\em Florence, Italy, 28 July--2 August
  2019}}, pages 1340--1350.

\bibitem[{Markert et~al.(2012)Markert, Hou, and Strube}]{markert12}
Katja Markert, Yufang Hou, and Michael Strube. 2012.
\newblock \href {http://www.aclweb.org/anthology/P12-1084.pdf} {Collective
  classification for fine-grained information status}.
\newblock In \emph{Proceedings of the 50th Annual Meeting of the Association
  for Computational Linguistics, {\em Jeju Island, Korea, 8--14 July 2012}},
  pages 795--804.

\bibitem[{McCann et~al.(2018)McCann, Keskar, Xiong, and
  Socher}]{McCann2018decaNLP}
Bryan McCann, Nitish~Shirish Keskar, Caiming Xiong, and Richard Socher. 2018.
\newblock The natural language decathlon: Multitask learning as question
  answering.
\newblock \emph{arXiv preprint arXiv:1806.08730}.

\bibitem[{Napoles et~al.(2012)Napoles, Gormley, and Durme}]{napoles12}
Courtney Napoles, Matthew Gormley, and Benjamin~Van Durme. 2012.
\newblock Annotated {G}igaword.
\newblock In \emph{Proceedings of the Joint Workshop on Automatic Knowledge
  Base Construction \& Web-scale Knowledge Extraction (AKBC-WEKEX) {\em
  Montr{\'{e}}al, Qu{\'{e}}bec, Canada, 7-8 June 2012}}, pages 95--100.

\bibitem[{Nedoluzhko et~al.(2009)Nedoluzhko, M{\'\i}rovsk{\`y}, and
  Pajas}]{nedoluzhko09}
Anna Nedoluzhko, Ji{\v{r}}{\'\i} M{\'\i}rovsk{\`y}, and Petr Pajas. 2009.
\newblock The coding scheme for annotating extended nominal coreference and
  bridging anaphora in the {P}rague dependency treebank.
\newblock In \emph{Proceedings of the Third Linguistic Annotation Workshop at
  ACL-IJCNLP 2009, {\em Suntec, Singapore, 6--7 August 2009}}, pages 108--111.

\bibitem[{Nissim et~al.(2004)Nissim, Dingare, Carletta, and
  Steedman}]{nissim04}
Malvina Nissim, Shipara Dingare, Jean Carletta, and Mark Steedman. 2004.
\newblock An annotation scheme for information status in dialogue.
\newblock In \emph{Proceedings of the 4th International Conference on Language
  Resources and Evaluation, {\em Lisbon, Portugal, 26--28 May 2004}}, pages
  1023--1026.

\bibitem[{Opitz and Frank(2018)}]{opitz-frank-2018}
Juri Opitz and Anette Frank. 2018.
\newblock Addressing the {W}inograd {S}chema {C}hallenge as a sequence ranking
  task.
\newblock In \emph{Proceedings of the First International Workshop on Language
  Cognition and Computational Models}, pages 41--52.

\bibitem[{Parker et~al.(2011)Parker, Graff, Kong, Chen, and
  Maeda}]{gigaword5.0data}
Robert Parker, David Graff, Junbo Kong, Ke~Chen, and Kazuaki Maeda. 2011.
\newblock English {G}igaword {F}ifth {E}dition.
\newblock LDC2011T07.

\bibitem[{Poesio(2004)}]{poesio04a}
Massimo Poesio. 2004.
\newblock The {M}{A}{T}{E}/{G}{N}{O}{M}{E} proposals for anaphoric annotation,
  revisited.
\newblock In \emph{Proceedings of the 5th SIGdial Workshop on Discourse and
  Dialogue, {\em Cambridge, Mass., 30 April -- 1 May 2004}}, pages 154--162.

\bibitem[{Poesio et~al.(2018)Poesio, Grishina, Kolhatkar, Moosavi, R{\"o}siger,
  Roussel, Simonjetz, Uma, Uryupina, Yu, and Zinsmeister}]{poesio18}
Massimo Poesio, Yulia Grishina, Varada Kolhatkar, Nafise~Sadat Moosavi, Ina
  R{\"o}siger, Adam Roussel, Fabian Simonjetz, Alexandra Uma, Olga Uryupina,
  Juntao Yu, and Heike Zinsmeister. 2018.
\newblock Anaphora resolution with the {ARRAU} corpus.
\newblock In \emph{Proceedings of the Workshop on Computational Models of
  Reference, Anaphora and Coreference. {\em New Orleans, Louisiana, June 6,
  2018}}, pages 11--22.

\bibitem[{Poesio et~al.(2004)Poesio, Mehta, Maroudas, and Hitzeman}]{poesio04d}
Massimo Poesio, Rahul Mehta, Axel Maroudas, and Janet Hitzeman. 2004.
\newblock Learning to resolve bridging references.
\newblock In \emph{Proceedings of the 42nd Annual Meeting of the Association
  for Computational Linguistics, {\em Barcelona, Spain, 21--26 July 2004}},
  pages 143--150.

\bibitem[{Poesio and Vieira(1998)}]{poesio98}
Massimo Poesio and Renata Vieira. 1998.
\newblock A corpus-based investigation of definite description use.
\newblock \emph{Computational Linguistics}, 24(2):183--216.

\bibitem[{Poesio et~al.(1997)Poesio, Vieira, and Teufel}]{poesio97b}
Massimo Poesio, Renata Vieira, and Simone Teufel. 1997.
\newblock Resolving bridging references in unrestricted text.
\newblock In \emph{Proceedings of the ACL Workshop on Operational Factors in
  Practical, Robust Anaphora Resolution for Unrestricted Text, Madrid, Spain,
  July 1997}, pages 1--6.

\bibitem[{Rahman and Ng(2012)}]{rahman12-winograd}
Altaf Rahman and Vincent Ng. 2012.
\newblock Resolving complex cases of definite pronouns: The {W}inograd {S}chema
  {C}hallenge.
\newblock In \emph{Proceedings of the 2012 Conference on Empirical Methods in
  Natural Language Processing and Natural Language Learning, {\em Jeju Island,
  Korea, 12--14 July 2012}}, pages 777--789.

\bibitem[{Rajpurkar et~al.(2016)Rajpurkar, Zhang, Lopyrev, and
  Liang}]{Rajpurkar16}
Pranav Rajpurkar, Jian Zhang, Konstantin Lopyrev, and Percy Liang. 2016.
\newblock {SQ}u{AD}: 100, 000+ questions for machine comprehension of text.
\newblock In \emph{Proceedings of the 2016 Conference on Empirical Methods in
  Natural Language Processing, {\em Austin, Texas, USA, 1--4 November 2016}},
  pages 2383--2392.

\bibitem[{R{\"o}siger(2018)}]{roesiger18b}
Ina R{\"o}siger. 2018.
\newblock {BASHI}: A corpus of wall street journal articles annotated with
  bridging links.
\newblock In \emph{Proceedings of the 11th International Conference on Language
  Resources and Evaluation, {\em Miyazaki, Japan, 7--12 May 2018}}, pages
  382--388.

\bibitem[{R{\"o}siger et~al.(2018)R{\"o}siger, Riester, and Kuhn}]{roesiger18a}
Ina R{\"o}siger, Arndt Riester, and Jonas Kuhn. 2018.
\newblock \href {http://aclweb.org/anthology/C18-1298} {Bridging resolution:
  Task definition, corpus resources and rule-based experiments}.
\newblock In \emph{Proceedings of the 27th International Conference on
  Computational Linguistics, {\em Santa Fe, New-Mexico, USA, 20--26 August
  2018}}, pages 3516--3528.

\bibitem[{Weischedel et~al.(2011)Weischedel, Palmer, Marcus, Hovy, Pradhan,
  Ramshaw, Xue, Taylor, Kaufman, Franchini, El-Bachouti, Belvin, and
  Houston}]{ontonotes4.0data}
Ralph Weischedel, Martha Palmer, Mitchell Marcus, Eduard Hovy, Sameer Pradhan,
  Lance Ramshaw, Nianwen Xue, Ann Taylor, Jeff Kaufman, Michelle Franchini,
  Mohammed El-Bachouti, Robert Belvin, and Ann Houston. 2011.
\newblock Onto{N}otes release 4.0.
\newblock LDC2011T03, Philadelphia, Penn.: Linguistic Data Consortium.

\bibitem[{Wu et~al.(2020)Wu, Wang, Yuan, Wu, and Li}]{wu-acl2020}
Wei Wu, Fei Wang, Arianna Yuan, Fei Wu, and Jiwei Li. 2020.
\newblock Coreference resolution as query-based span prediction.
\newblock In \emph{Proceedings of the 58th Annual Meeting of the Association
  for Computational Linguistics, {\em Seattle, Wash., 5--10 July 2020}}.

\end{thebibliography}
\end{document}